\documentclass[dvips,aos]{imsart}
\RequirePackage[colorlinks,citecolor=blue,urlcolor=blue]{hyperref}

\usepackage{graphicx} % Include figure files
\usepackage{dcolumn}
\usepackage{amsmath}
\usepackage{amsfonts}
\usepackage{bm}

\startlocaldefs
\newcommand{\func}[3]{\ensuremath{\operatorname{#1}_{#2}\left(#3\right)}}
\newcommand{\efunc}[1]{\ensuremath{\mathrm{e}^{#1}}}
\newcommand{\floor}[1]{\ensuremath{\left\lfloor#1\right\rfloor}}
\newcommand{\ceil}[1]{\ensuremath{\left\lceil#1\right\rceil}}
\newcommand{\abs}[1]{\ensuremath{\left|#1\right|}}
\newcommand{\conj}[1]{\ensuremath{\bar #1}}
\newcommand{\openone}{\ensuremath{\mathbf 1}}
\endlocaldefs

\begin{document}

\begin{frontmatter}
% "Title of the paper"
\title{Resolution and Scale Independent Function Matching Using a String       Energy Penalized Spline Prior}
%\runtitle{Scale Independent Function Matching}
\runtitle{Res. and Scale Indep. Fn. Matching}

% indicate corresponding author with \corref{}
% \author{\fnms{John} \snm{Smith}\corref{}\ead[label=e1]{smith@foo.com}\thanksref{t1}}
% \thankstext{t1}{Thanks to somebody} 
% \address{line 1\\ line 2\\ printead{e1}}
% \affiliation{Some University}

\begin{aug}
\author{\fnms{David M.} \snm{Rogers}\ead[label=e1]{rogersdd@uc.edu}\thanksref{a1,a2,m1}}
%\address{\printead{e1}}
\and
\author{\fnms{Thomas L.} \snm{Beck}\ead[label=e2]{thomas.beck@uc.edu}\thanksref{a2,m1,m2}}
\affiliation{University of Cincinnati}
\address{Departments of Chemistry\thanksmark{m1} and Physics\thanksmark{m2}\\
University of Cincinnati\\
Cincinnati, OH 45221-0172\\
\printead{e1}\\
\phantom{E-mail:\ }\printead*{e2}}

\thankstext{a1}{Supported by a DOE Computational Science Graduate Fellowship DE-FG02-97ER25308.}
\thankstext{a2}{Supported by Army MURI Grant DAAD19-02-1-0227 and NSF grant CHE-0709560.}

\runauthor{D. M. Rogers and T. L. Beck}
\end{aug}

\begin{abstract}

  The extension of the Bayesian penalized spline method to inference on vector-valued functions is considered, with an emphasis on characterizing the suitability of the method for general application.
We show that the standard quadratic penalty is exactly analogous to the energy of a stretched string, with the penalty parameter corresponding to its tension.  This physical analogy motivates a discussion of resolution independence, which we define as the convergence of our function estimate to the nonparametric solution as the spline width decreases.
The multidimensional context makes direct application of standard procedures for choosing the penalty parameter difficult, and a new method is proposed and extensively compared to the existing literature.
An important class of problems which can be analyzed by this method are stochastic numerical integrators, which are considered as an example problem.  This work represents the first extension of penalized spline methods to inference on multidimensional numerical integrators reported in the literature.
Several numerical calculations illustrate the above points and address practical application issues.

\end{abstract}

% + 62G35 Robustness (nonparametric inference)
% + 70F17 Inverse problems (in particle dynamics)
%   65D10 Smoothing, curve fitting
%   62G08 Nonparametric regression
%   41A15 Spline approximation

% considered, not used:
%   65C40 Computational Markov chains
%   65D07 Splines
%   65D15 Algorithms for functional approximation
%   70G70 Functional-analytic methods (in particle dynamics)
%   82C22 Interacting particle systems

\begin{keyword}[class=AMS]
\kwd[Primary ]{62G35}
\kwd[; secondary ]{70F17}
\kwd{65D10}
\kwd{62G08}
\kwd{41A15.}
\end{keyword}

\begin{keyword}
\kwd{Multidimensional}
\kwd{nonparametric regression}
\kwd{scale invariant}
\kwd{smoothing}
\kwd{spline.}
\end{keyword}

\end{frontmatter}

% make the title area
%\maketitle

\section{ Introduction}
\label{sec:intro}

\subsection{ Motivation and Problem Background}

  Function estimation from observational data is an important modeling tool because it allows us to explore the relationship between the independent variable and its response variable as well as predict observations yet to be made.  Some simple example applications involve estimating continuous changes in a property with respect to time (e.g. helmet acceleration during impact \cite{bsilv85} or dose-response curves \cite{ajull07}).  More advanced applications can involve more than one dimensional input, for example estimating scalar functions of space and/or time variables \cite{slang04,abrez06}.  Here the function is modeled as a linear combination of functions with one or two-dimensional arguments, giving it the classification of a generalized additive model~\cite{thast90}.  Yet another generalization is possible when the function is vector-valued~\cite{tyee96}, for example the force on a particle moving in three dimensions \cite{ferco94}.  We find in this last case the framework for analyzing a large class of novel applications such as stochastic integrators.

  In order for a function estimation method to be robust with respect to the experimental setup, it must satisfy two major requirements.  First, the estimator must be resolution independent in the sense that it becomes stable in the limit of arbitrarily high resolution (where an acceptable approximation to true function is almost certainly in the parameter space).  Standard least squares estimators (maximum likelihood estimation using Eq.~\eqref{eq:form} with $\lambda = \vec 0$ and without $\log \lambda_{ti}$ terms)\cite{NRgls} do not satisfy this criterion.
Smoothing splines were shown to have this property~\cite{crein67} and even an asymptotic equivalence to convolution kernel-based smoothing in the early work of B. Silverman~\cite{bsilv84,bsilv85}, which was carried on by K. Messer~\cite{kmess91} and D. Nychka~\cite{dnych95}.
A subsequent improvement in the original formulation came through divorcing the measurement positions, $\{r^l\}$, from the spline knots~\cite{peile96} which allows for more efficient calculations.
The resolution independence of penalized splines allows us to overcome the difficult ``bin-width" problem encountered in linear least-squares fitting, where too low or too high resolution cause unphysically smooth curves or overfitting.
Second, the estimator must be scale independent in the sense that it predicts the correct function shape when both the function and its noise are subject to arbitrary scaling.  The Akaike information criterion (AIC) method fails this test, while generalized cross-validation (GCV) is scale independent but fails to separately address estimation of the sample variance.

  The physical spline device already satisfies the infinite resolution limit sought above for a nonparametric method.  It consisted of a section of rubber held in place with push-pins at several points along a desired smooth curve \cite{rwilk57}.  This device could be numerically implemented by simply minimizing the potential energy function for (a suitable computational description of) a string's position subject to the given constraint points.
However, modern applications of function fitting are often based on noisy measurements so that we cannot assume the input data $\{y,r\}_1^M$ to constitute absolute constraints.  To modify the above device for this situation, imagine placing a push-pin at the location of each input measurement and making its connection to the spline by attaching a vertical spring.  This is, in fact, the system we are led to by the Bayesian analysis of Sec.~\ref{sec:deriv} -- where the role of the spring constants are played by $z$, the inverse of the measurement variance.
In addition, the specification of an energy function leads to a complete description of the probability density for our spline's position if Boltzmann statistics are assumed.  This assumption equates the minimum energy solution to the popular maximum likelihood estimate of statistics.% \cite{}

  Unfortunately, the system's potential energy function now depends jointly on the spring constants and the spline tension, $\lambda$.  This causes the ratio between the two ($\lambda/z$) to decide the form of the final solution.  Nevertheless, it is an important result of this report that when proper prior probabilities are chosen for the tension and spring constants, scale-independence is achieved.

  In order to motivate the choice of B-spline basis functions used to describe $f(r)$, we note some of their properties here.  In keeping with the nonparametric requirement we require that the basis set be able to approximate any sufficiently smooth function to good accuracy.
Using B-spline basis functions to represent the function space of interest allows a choice of the function's continuity, via the spline order, as well as its resolution, via the number of free parameters or knots.  These choices can be important in numerical applications of the fitted function.  In addition, for all choices of the knot spacing and spline order, the best B-spline approximation can be shown to have the same approximation error as a polynomial fitting~\cite{ureif97}.

  Several issues in using penalized splines have been addressed by previous studies.  In attempts to understand scale dependence, several proposals for choosing the penalty scale parameter have been considered~\cite{jrice84,slang04,ajull07}, including introduction of non-uniform (in $r$) penalty and/or variance parameters~\cite{jfrie91,cbill00,drupp00,vbala05}.  Adaptation to cases with non-Gaussian sampling error have been addressed by consideration of inference schemes with alternative likelihood functions~\cite{maert02}.  Notably, several authors have interpreted the complete fitting process as a Bayesian inference problem~\cite{slang04}.  This leads to consideration of the penalty function as a prior distribution on function space, with the attendant penalty parameter as a (nuisance) hyperparameter.

\subsection{ Material Covered}

  In view of the above considerations, it seems that each new application area of P-splines should reconsider the proper choice of prior probability for function space and likelihood function for the problem at hand.  This report will attempt to address these issues in the Bayesian formulation of P-splines, so that the method can be used in new applications with confidence.

  First, section~\ref{sec:deriv} presents a physically motivated re-derivation of the prior penalty by likening the function to a stretched string.  The shortcomings of the standard, scale-invariant prior probability for the penalty parameter are re-examined and found to be due to the possibility of a degenerate polynomial solution.  We show that we can exclude this possibility by introducing a string zero-point energy to modify the standard Jeffrey's prior, while retaining scale invariance for arbitrarily large measurement scalings.

  Next we derive important properties of the proposed prior distribution in section~\ref{sec:props}.  Resolution independence is proved by finding that the approximation error of the $n^\text{th}$ derivative of a B-spline solution is of order $h^{r-n}$.  A review of the link between spline and convolution-based smoothing shows the role of higher-order derivatives in the penalty function.  The marginal distribution of the penalty parameter serves to further clarify the role of the zero-point energy and greatly simplify sensitivity analysis to this fixed numerical value.
The insight gained in these two sections allows automatic identification of the scale parameter in new application areas.

  Section~\ref{sec:probs} compares results from Markov chain Monte-Carlo (MCMC) simulations on several test applications to provide a numerical test of the scale independence sought.  An implementation of the multi-dimensional method for matching the output of a stochastic integrator is also reviewed and tested.  For the particular problem studied, it is found that use of the posterior average estimator gives significantly better results than maximum likelihood estimation for small sample sizes.

  Finally, notes on B-splines and computational implementation of the integrations necessary for calculating higher order derivatives in the penalty function are provided in the appendix.

\subsection{ Problem Formulation}

  We consider the regression problem where $M$ (possibly noisy) measured values, $\{y^l,r^l\}_1^M$, of a property $y = f(r)+\text{noise}$ are to be fitted to an arbitrary function $f(r) = \sum_{k=1}^{N_f} B_k(t_k(r)) = \mathbf D(r) \cdot \theta$.  We allow for the possibility that $y$ and $r$ are vector-valued by choosing the design matrix $\mathbf D(r) \in \mathbb M_{N \times p}$.  In order to infer the value of the parameters $\theta \in \mathbb R^p$, we derive the penalty function
\begin{equation}
\label{eq:form}
\begin{split}
-\log &\func{P}{}{\theta \lambda z | \{y^l,r^l\}_1^M I} = \text{const.} \\
  & + \sum_{i=1}^N \left\{ \tfrac{z_i}{2} \left[\sum_{l=1}^M (y^l_i - f_i(r^l))^2+V_0\right]
 - \left(\tfrac{M}{2}-1\right) \log z_i \right\} \\
  & + \sum_{k=1}^{N_f} \left\{ \tfrac{\lambda_k}{2} \left[\int_{{\mathbb S}_{t_k}}{B_k^{(n)}(t_k(r))^2 \rho(t_k) dt_k} + E_0\right]
 - \left(\tfrac{p_k-n}{2}-1\right) \log \lambda_k \right\}
.
\end{split}
\end{equation}

  The indices are necessarily introduced by allowing each dimension of $y$ to have its own measurement variance $\sigma^2_i = z^{-1}_i$, and $f(r)$ to be composed of the sum of $N_f$ B-spline functions $B(t)$ of one-dimensional variables $t_k(r) \in {\mathbb S}_{t_k}$ and parameters $\theta_k$ such that $\sum_{k=1}^{N_f} p_k = p$.  Each of these functions has its own roughness penalty of order $n$, whose scale parameter (or tension) is $\lambda_k$.  We can analyze this complicated form by considering the simple case where $N$ and $N_f$ are both one, making $t_1(r) = r$ and
\begin{equation}
\label{eq:sform}
\begin{split}
-\log &\func{P}{}{\theta \lambda z | \{y^l,r^l\}_1^M I} = \text{const.} \\
  & + \tfrac{z}{2} \left[\sum_{l=1}^M (y^l - B(r^l))^2+V_0\right]
 - \left(\tfrac{M}{2}-1\right) \log z \\
  & + \tfrac{\lambda}{2} \left[\int_{{\mathbb S}_{r}}{B^{(n)}(r)^2 \rho(r) dr} + E_0\right]
 - \left(\tfrac{p-n}{2}-1\right) \log \lambda
.
\end{split}
\end{equation}

\section{ Derivation of the Method}
\label{sec:deriv}

\subsection{ Thermal Equilibrium of a Stretched String}

  The prior probability for the space of all possible functions has been widely taken to be a Gaussian distribution on the function's spline parameters.  The penalty matrix (inverse of the variance-covariance) is either an $n^\text{th}$ order difference on successive spline parameters or derived from an integral over the squared $n^\text{th}$ order derivative of the function -– corresponding to an $n^\text{th}$ order random walk of the control points or the function $f(r)$, respectively~\cite{peile96,slang04}.

  Here, an appeal to the latter, traditional, penalty function is laid out based on consideration of the spline fit as a smooth string.  In the tradition of the classical spline device\cite{rwilk57}, it seems appropriate to consider the function's $n-1^\text{th}$ derivative $f^{(n-1)}(r)$ as such a stretched string with tension $T(r)$.  Once a suitable energy function for the string is available, the assumption of Boltzmann statistics specifies a probability distribution for the string's position (and hence its parameters).

% Dirichlet, periodic, or const. slope??
  Under Dirichlet boundary conditions, the standard expression for a string's potential energy is given by [using the notation from Eq.~\eqref{eq:sform}]:
\begin{equation}
\label{eq:stren}
\operatorname{E}\left[f(r)\right] = \frac{1}{2} \int T(r) {\left \| \nabla^{(n)} f(r) \right \|}^2 \rho(r) dr
\end{equation}
where $\rho(r) dr$ is the volume element for integration (i.e. $dr$ for a line, $4 \pi r^2 dr$ for a spherically symmetric membrane, etc.).  For uniform tension $T_0$ (or tension with unknown scale but known variation with respect to $r$) and a B-spline basis for $f(r)$, this formula can be integrated to give the familiar P-spline penalty function:
\begin{equation}
\label{eq:pen}
\operatorname{E}\left[f(r)\right]/V = \frac{T_0}{2} \theta^T \cdot \mathbf Q \cdot \theta
\end{equation}
where we have made the definitions:
\begin{align}
\label{eq:nderiv}
\nabla^{(n)}f(r) &= \mathbf A(r) \cdot \theta \\
\label{eq:Q}
\mathbf Q &= \frac{1}{V} \int \frac{T(r)}{T_0}
			{\mathbf A(r)}^T \cdot \mathbf A(r) \rho(r) dr \\
\label{eq:V}
V &= \int \rho(r) dr
\end{align}
The relation for the derivatives of $f(r)$, Eq.~\eqref{eq:nderiv}, follows from the linearity of B-splined functions in their parameters, $\theta$.
As we allow a separate $\mathbf Q$ for each additive function of one variable, $\mathbf A$ is simply a row-vector.  In this report, we distinguish matrix-multiplication from scalar multiplication with the symbol ``$\cdot$".

	%Where Q is identified as the weighted integral of the square of the first derivative times a factor which cancels its x-units (spline height h2n divided by the total integration volume, V).  The penalty matrix, Q, of this derivation is therefore equivalent to h/V times that of an nth order random walk when the spline order is n+1.  This can be seen by noting that the difference of B-spline parameters (times du/dx = h-1) gives new B-spline coefficients for the function derivative.  Recursively applying this property gives an order 1 B-spline, whose squared integral is just the sum of its squared parameters (times h).  Note also that the integration interval matches in the case of n+1th order nonperiodic B-splines defined on (n,N) and periodic B-splines with periodic difference operators.

  Likening the penalty to the spline energy gives the following connection to statistical mechanics.  A system allowed to exchange heat with its surroundings until thermal equilibrium is reached has a Boltzmann probability distribution:
\begin{equation}
\label{eq:boltzmann}
\func{P}{}{x} dx \propto \efunc{-\beta \func{E}{}{x}} dx
\end{equation}
Where the inverse thermal energy $\beta$ determines the scale of the energy fluctuations (and thus the spline roughness) allowed in the system, whose position is specified by coordinates $x$.  Defining the product $\beta V T_0$ as $\lambda$ recovers the traditional penalty function and assigns the meaning of ``dimensionless tension" to the penalty parameter, $\lambda$.

  Applying Boltzmann statistics to our string system thus gives us an improper normal distribution, since the $n$ string modes corresponding to coefficients of an $n-1$ degree polynomial have been assigned a uniform prior due to their absence from the energy function.
\begin{equation}
\label{eq:ptheta}
\func{P}{}{\theta | \lambda I } \propto \lambda^{(p-n)/2} \efunc{-\tfrac{\lambda}{2} \theta^T \cdot \mathbf Q \cdot \theta}
\end{equation}

  It is interesting to note that the above formula starts from a string energy density which is dependent on the shape of the function, not the number of spline parameters, $p$.  However, the prior becomes dependent on $p$ through the rank of $\mathbf Q$ when normalized by integration over $\theta$.  From a mechanical perspective, this occurs simply because larger numbers of parameters allow the function much more free space to move.  This makes the probability of any individual choice of $\theta$ correspondingly less.  From a Bayesian perspective, this can be justified by considering how the number of functions within a small region of function space grows with $p$.  We have thus recovered the common element of all previous formulations of P-splines via invoking Boltzmann statistics on an intuitive physical model system.

\subsection{ Penalty and Variance Prior Distributions}
\label{sec:pprior}

  As noted in the introduction, however, the penalty parameter and sample variance remain to be dealt with.  The standard Bayesian choice is a simple scale-invariant prior for $\lambda$, giving Eq.~\ref{eq:prior} with $E_0=0$ as the complete prior.
Although this approach works most of the time, it has one serious numerical issue which appears in the limit as $\lambda$ approaches infinity.  This situation occurs when the sample size is small and the derivative order is large, allowing the belief that an $n-1$ degree polynomial is a possible solution to the inference problem (i.e. a singularity at $\lambda=\infty$).  Because MCMC sampling alternates between drawing values for $\theta$ and $\lambda$, it can get stuck once the algorithm encounters a $\lambda$ large enough to force a polynomial solution.

  Because of the divergence at large $\lambda$, this parameter requires an alternative prior distribution as has been done by several authors.  Indeed, the earlier non-Bayesian suggestions for penalty parameter selection implicitly suggest a prior on $\lambda$.  Our analogy to stretched strings provides an alternative method for deriving such a prior.  We gain a vital clue by the above noted degeneration of the sampling process into a least-squares polynomial fit. Since our problem formulation in terms of stochastic splines implies an aversion to such simple solutions, we need to find a way to explicitly add this aversion as prior information.  Adding $\nu$ observations of small displacements, ${\left \| \nabla^{(n)} f(r_i) \right \|}^2 = w_i^2$ to our state of knowledge gives a Gamma distribution for $\lambda | \left(E_0 = \sum_1^\nu w_i^2 \right)$ with parameters $a_0 = \nu/2$, $b_0 = E_0/2$.  This approach was tried by Lang and Brezger~\cite{slang04} with $\nu=2$ and found to give results dependent on the scale of the function measurements due to bias toward $a_0/b_0$ -- which is the prediction for $\lambda$ implied by the displacements assumed in the new prior.  This prediction for $\lambda$ is effectively specifying an energy scale for measuring displacements in our string.  Jullion and Lambert~\cite{ajull07} propose to set $\nu=2$ and provide a hyperprior for $E_0/\nu$.  The corresponding interpretation is that $\nu$ prior observations of the string displacement were made but subject to large uncertainty.

  However, these approaches assume some prior knowledge about both ends of the energy scale, where the original intent of the additional information was simply to eliminate numerical instability at the high tension (degenerate solution) end.  The usual difficulty with scale-invariant priors is divergence toward zero due to lack of observations of the corresponding process' scale.  Here, observations of the tension parameter are already made through the movements of the function away from a polynomial form.  Divergence toward infinity as noted above should only occur if the input points actually lie on a polynomial and no noise occurs in the system.  In all other cases, such divergence is an unwanted numerical artifact.  In order to deal with this artifact, suppose that we arbitrarily set a value for $\theta^T \cdot \mathbf Q \cdot \theta$ at which we consider the spline to be a degenerate polynomial solution.  Since we are not interested in variations of the function below this threshold, we propose modifying the standard scale-invariant prior to impose this condition on the sampling process by adding $E_0$ to $\theta^T \cdot \mathbf Q \cdot \theta$.
As noted in the discussion above, this is equivalent to the limit of specifying a string containing a zero-point energy, $E_0$, without any explicit observations (i.e. $\nu=0$).
This argument shows that $E_0$ is a measure of residual uncertainty that the solution is an $n-1$ degree polynomial.

\begin{equation}
\begin{split}
\label{eq:prior}
\func{P}{}{\theta \lambda | I} &= \func{P}{}{\theta | \lambda I}
                                  \func{P}{}{\lambda | I} \\
  &\propto \lambda^{(p-n)/2-1} \efunc{ -\tfrac{\lambda}{2}
          \left( \theta^T \cdot \mathbf Q \cdot \theta + E_0 \right) }
\end{split}
\end{equation}

  Setting $\nu=0$ has important consequences for previously reported problems with this prior~\cite{ajull07}, since the implied prior distribution for $\lambda$ now approximates the scale independent prior, in particular its cumulants approach zero independent from $E_0$.  In fact, the new prior distribution on $\ln \lambda$ is sigmoid-like, going from $1$ at $-\infty$ to $1/2$ at $\ln \lambda_{1/2} = \ln(2 \ln 2) - \ln E_0$ (with slope $-(\ln 2)/2$) to $0$ at $\infty$.

\subsection{Bayesian Posterior Parameter Distribution}
\label{sec:post}

  Assuming independent Gaussian likelihood functions with variance $\sigma^2 \equiv 1/z$, the posterior distribution is given by
\begin{equation}
\begin{split}
\label{eq:post}
\func{P}{}{\theta z \lambda | D I} &\propto \lambda^{(p-n)/2-1} z^{M/2-1} \\
  & \times \exp \left\{
  -\tfrac{z}{2} \left( \left\|Y-\mathbf D \cdot \theta\right\|^2 + V_0 \right)
  -\tfrac{\lambda}{2} \left( \theta^T \cdot \mathbf Q \cdot \theta + E_0 \right)
       \right\}
.
\end{split}
\end{equation}
And the conditional posterior distributions are therefore
\begin{align}
\label{eq:postth}
\theta | \cdots &\sim N(\bar \theta, \mathbf \Sigma) \\
		&\mathbf \Sigma^{-1} = \lambda \mathbf Q + z {\mathbf D}^T \cdot \mathbf D, \qquad
		  {\mathbf \Sigma}^{-1} \cdot \bar \theta = z {\mathbf D}^T \cdot Y \notag \\
\label{eq:postz}
z | \cdots &\sim \Gamma \left(M/2, (\|\mathbf D \cdot \theta - Y\|^2+V_0)/2\right) \\
\label{eq:postl}
\lambda | \cdots &\sim \Gamma \left((p-n)/2, (\theta^T \cdot \mathbf Q \cdot \theta+E_0)/2 \right)
,
\end{align}
where we have formed a matrix $\mathbf D \in {\mathbb M}_{M \times p}$ by stacking row-vectors of spline coefficients $\mathbf B(r)$ from all observations.  Similarly, $Y$ is a column-vector of all the observed function values.

  An additional parameter appears because the same type of situation described for $\lambda$ can also occur for the inverse variance parameter when the number of measured data points is low compared to the spline resolution.  In this case, the spline can exactly match the input measurements resulting in a singularity as $z \rightarrow \infty$.  This case has been studied in detail by Wahba and Wang~\cite{gwahb95}.  Our remedy will be just as above, adding a minimum variance $V_0$ into the residual sum of squares.

  Using MCMC techniques~\cite{rrubi81,dgame06} to sample the posterior distribution Eq.~(\ref{eq:postth}--\ref{eq:postl}) is the direct analogue of observing the multiple positions of a spline device (as described in the introduction) in thermal equilibrium.
Alternately, a maximum likelihood estimate can be carried out using relatively fewer iterations of the above steps by simply maximizing each conditional posterior and iterating until convergence.
Both processes require a Cholesky decomposition for $\theta$ at each iteration, which is an ${\mathbb O}(p^3)$ operation.

	We can calculate the classical mean-squared error (MSE) of the function estimate $\hat f(r)$ using Eq.~\ref{eq:postth} to compare with standard results~\cite{dnych90}.  This procedure gives an indication of the convergence toward a known $\theta_0$.  This is in contrast to the Bayesian {\it a posteriori} error estimate for $\bar \theta$, which is given by the covariance matrix $(z {\mathbf D}^T \cdot \mathbf D + \lambda \mathbf Q)^{-1} \equiv \mathbf K / (z M)$. The squared error is expressed as $M^{-1} \left\| Y_0 - \hat Y \right\|^2 \approx \int{(f(r)-\hat f(r))^2 dr}$ and its expectation is taken with respect to all possible variations of the random noise.  For the one-dimensional case, we therefore assume the vector of observations is $Y = \mathbf D \cdot \theta_0 + \epsilon$, and the corresponding estimator and error PDF are:
\begin{align}
\label{yhat}
\hat Y &= \mathbf D \cdot \mathbf K \cdot {\mathbf D}^T \cdot (\mathbf D \cdot \theta_0/M + z^{-1/2} \epsilon/M) \\
Y_0 - \hat Y | \theta_0 \{r\}_1^M \lambda z I &\sim N \left(\mathbf D \cdot \mathbf K \cdot \tfrac{\lambda}{z M} \mathbf Q \cdot \theta_0,
	\tfrac{1}{z M^2} (\mathbf D \cdot \mathbf K \cdot {\mathbf D}^T)^2 \right)
\end{align}
Where we have used $\theta_0 = \mathbf K \cdot ({\mathbf D}^T \cdot \mathbf D/M + \tfrac{\lambda}{z M} \mathbf Q) \cdot \theta_0$ and $\epsilon$ is a vector of $M$ standard normal random variables.
The expected error is then
\begin{equation}
\label{eq:MSE}
\text{MSE} = M^{-1} \left\{ M^{-2} \left\|\mathbf D \cdot \mathbf K \cdot \tfrac{\lambda}{z} \mathbf Q \cdot \theta_0 \right\|^2
 + z^{-1} \mathrm{Tr}\left[ (M^{-1} \mathbf D \cdot \mathbf K \cdot {\mathbf D}^T )^2 \right] \right\}
.
\end{equation}

  The second term on the right hand side is reminiscent of the Bayesian estimate, while the first term represents the bias introduced by assuming function smoothness.  For a fixed $\lambda/z$, the total dependence on $M$ for this term (remembering $\mathrm{Tr}(\mathbf D^T\cdot \mathbf D)=M$ for periodic splines) is order $M^{-2}$.  However, this is not a direct estimate of the bias of the full Bayesian inference scheme, which allows $\lambda/z$ to vary.

\subsection{ Multi-dimensional generalization}
\label{sec:multidim}
  We first generalize Eq.~\eqref{eq:postth} to the case where $f(r) = \sum_{k=1}^{N_f} f_k(r)$ is the sum of multiple additive scalar functions.  We can do this by extending the basis for $y(r) = {\mathbf D}(r) \cdot \theta$ by tacking on additional coefficients.  For the case of $N_f$ functions $f_k(r) = \mathbf B_k(r) \cdot \theta_k$ with spline coefficients ${\mathbf B}_k(r)^T \in \mathbb R^{p_k}$, this means that ${\mathbf D}(r) = [{\mathbf B}_1(r), {\mathbf B}_2(r), \ldots, {\mathbf B}_{N_f}(r)]$ -- taking the total number of parameters to be $p = \sum_{k=1}^{N_f} p_k$.  Each additive function should have its own string energy prior, making $\lambda$ a vector of dimension $N_f$ and $\mathbf Q$ a set of matrices penalizing their respective functions.
The posterior probability for each element of $\lambda$ replacing Eq.~\eqref{eq:postl} is thus
\begin{equation}
\label{eq:postlm}
\lambda_k | \cdots \sim \Gamma \left((p_k-n)/2, (\epsilon_{Q_k}^2+E_0)/2 \right)
.
\end{equation}
In addition, when $N_f > 1$, each function can only be determined to within an additive constant, and identifiability constraints must be placed on all but one of them. %ref?

  Next, we generalize to vector-valued functions for models of the form $f(r_l) = \mathbf D_l \cdot \theta$, where $\mathbf D_l$ is an $N \times p$ matrix.  This can be constructed by summing matrices formed from the tensor product of any direction vector $g_k(r) \in \mathbb R^N$ with the familiar spline coefficients:
\begin{equation}
f_k(r) = g_k(r) \otimes {\mathbf B}_k(r) \cdot \theta_k
\end{equation}

  Identifiability problems are worse for the multidimensional case.  For definiteness, assume without loss of generality that each function has a unique argument $B_k(r)=B_k(t_k(r))$.  If two functions $B_{k1}$ and $B_{k2}$ have the same argument but different directions, they can combined to make a single unique function $B_k(r_k) = (g_{k1}(r)+g_{k2}(r)) \otimes {\mathbf B}(r_k) \cdot \theta_k$.  Cases with the same argument but different ranges are irrelevant, since their ranges can be combined to make a single, larger spline function or considered as two independent variables for the present analysis.  If, on the other hand, two functions with different arguments share the same directional vector for all samples encountered, an identifiability problem occurs.  We therefore make the further assumption that the sample size is sufficiently large so that if two directional vectors differ in some region in the space of $r$, that region is included in the sample.  Allowing multiple functions to share the same set of spline parameters collapses the total set of functions to identify via combining functions $B_k$ which share parameters $\theta_K$ to make a total of $N_F$ unique functions.  Note the use of capitalized indices for combined sets.
\begin{equation}
\label{eq:multfn}
\mathbf D_{l,K} = \sum_{k \in K} g_k(r^l) \otimes {\mathbf B}_k(r^l_k)
\end{equation}
This leaves us with genuine identifiability constraints which can be seen by examining
\begin{equation}
f(r^l) = \sum_{K=1}^{N_F} \mathbf D_{l,K} \cdot (\tilde \theta_K+C_K)
.
\end{equation}
Wherein $\tilde \theta_K$ denotes the constrained $\theta_K$ (i.e. its average has been subtracted) and $C_K$ are arbitrary additive constants.  The effect of constraining $\theta_K$ during sampling is to force $C_K$ to zero.  Any direction in which $\{C\}$ could vary while leaving $f(r)$ unchanged for every $r$ must therefore have a corresponding constraint.  The number of constraints is thus determined by the rank of $\mathbf R \in \mathbb M_{M \times N_f}$, formed by the column-vectors $\mathbf D_{*,K} \cdot \openone$ (associating $K$ with the column).  Using the definition \eqref{eq:multfn}, $\mathbf R_{l,K} = \sum_{k \in K} g_k(r_l) ({\mathbf B}_k(r_l) \cdot \openone) \equiv g_K(r_l)$, the assumptions above imply that the number of constraints is equal to the number of persistent linear dependencies among $g_K(r_*)$.

  It is worthwhile to notice that a multidimensional basis function can be constructed from differentiation of a scalar function of the familiar linear mixed model type (e.g. $f(r) = \sum \vec{\tfrac{\partial t_K(r)}{\partial r}} f_k(t_k)$).  When such derivative information is used, more data points than a corresponding sample from the scalar function are acquired with each sample.  In this case identifiability problems only occur when the function can be simplified by reducing the dimension of $\{t_K(r)\}_1^{N_F}$ in a linear algebraic way.

  Using a multivariate Normal likelihood function and assigning each dimension of the observation its own independent variance again gives a set of linear equations to solve during sampling, replacing the mean and variance of Eq.~\eqref{eq:postth} with
\begin{equation}
\label{eq:postthm}
{\mathbf \Sigma}^{-1} = \func{diag}{}{\lambda} \cdot \mathbf Q + \sum_{i=1}^N z_i {\mathbf D}_{*,i}^T \cdot \mathbf D_{*,i}, \qquad
                  {\mathbf \Sigma}^{-1} \bar \theta = \sum_{i=1}^N z_i {\mathbf D}_{*,i}^T Y_{*,i}
\end{equation}.
Where we have formed a list of matrices $\mathbf D \in {\mathbb M}_{S \times N \times p}$ by stacking $\mathbf D_l \in \mathbb M_{N \times p}$ for all observations and $\mathbf Q$ is understood to be an appropriate block-diagonal matrix of $\mathbf Q_k$-s.
Correlations between elements of each measurement can be incorporated by a trivial modification as long as they remain a known function of $r$.

If $N_I$ dimensions share a common $z_I$, the posterior probability for each element of $z$ replacing Eq.~\eqref{eq:postz} is
\begin{equation}
\label{eq:postzm}
z_I | \cdots \sim \Gamma \left(N_I M/2, (V_0+\sum_{i \in I}^{N_I} \|\mathbf D_{*,i} \theta - Y_{*,i}\|^2)/2\right)
\end{equation}.

\section{ Properties of the Proposed String Energy Prior}
\label{sec:props}

  Most of the properties of penalized splines can be analyzed analytically for a fixed ratio, $\alpha \equiv \tfrac{\lambda}{z}$.  This is due to the existence of unique solutions to the one-dimensional spline fitting problem for an unrestricted function in a Hilbert space~\cite{gkime70,crein67}.  In addition, Silverman~\cite{bsilv84} showed how the solution process can be understood in terms of a convolution kernel at the large sample size limit, establishing an analogy between convolution smoothing methods and spline smoothing~\cite{kmess91,kmess93,dnych95,fabra99}.  These results allow us to investigate resolution independence and the effects of differing penalty function orders, $n$.  To show how this result comes about and analyze its limitations, we will sketch a short and intuitive derivation here, while noting that more extensive research on this analogy has been carried out by others.

  Next, in order to expand these results for variable $\alpha$, we derive the marginal likelihood of the variance and penalty parameters.  This allows us to understand the properties of a Bayes' estimate for $\theta$ in terms of averages over $\alpha$ as well as study the trade-off between under or over smoothing.

\subsection{ Resolution Dependence of P-Spline Estimation}
\label{sec:res}

  A general formula for the Fourier transform of the convolution kernel [Eq.~\eqref{eq:tG}] corresponding to spline smoothing with arbitrary derivative order, $n$, was first given by Silverman~\cite{bsilv84}.  Subsequently, several authors improved the results on asymptotic convergence of smoothing spline and convolution-based methods, including Messer~\cite{kmess91} and Messer and Goldstein~\cite{kmess93}, where convenient approximate convolution kernels were derived for fixed sample spacing.  Nychka~\cite{dnych95} relaxed the equal spacing restriction and showed that the convolution kernel decays exponentially as long as the samples are spaced closely enough.  Finally, Abramovich and Grinshtein~\cite{fabra99} provided a systematic derivation of an asymptotically equivalent convolution kernel for arbitrary derivative order, sample spacing, and variable tension.  We will briefly sketch the results of the latter here, while providing the connection to finite element solutions as proposed here in order to establish resolution-independence.

  We begin with the task of finding the function $u \in {\mathbb H}^n(\Omega)$ (where ${\mathbb H}^n(\Omega)$ is the standard Sobolev space of functions with square integrable derivatives on $\Omega$ up to $n^{\text{th}}$ order) which minimizes the functional
\begin{equation}
\label{eq:Phi}
\Phi[u] = \tfrac{1}{M} \sum_{l=1}^M (y_l - u(t_l))^2 + \tfrac{\alpha}{V M} \int_{\Omega}{\rho(t) u^{(n)}(t)^2 dt}
.
\end{equation}
In the following discussion we are considering both the infinite, $\Omega = \mathbb R$, and periodic domains, $\Omega = [0,V)$.
Next, note that we can re-formulate the sum appearing on the right hand side as an integral using the definitions:
\begin{align}
y(t) &= \begin{cases}
\tfrac{\sum_{\{l:t = t_l\}} y_l}{\sum_{\{l:t = t_l\}} 1}, & t \in \{t_l\}_1^M\\
0, & \text{otherwise}
\end{cases} \notag \\
\phi(t) &= \tfrac{1}{M} \sum_{l=1}^M \delta(t-t_l)
,
\end{align}
so that $\Phi$ can be conveniently expressed as a norm using the scalar product $<a,b> = \int_\Omega \conj{a(t)} b(t) dt$, where $\conj{a}$ denotes complex conjugation.
\begin{equation}
\label{eq:phi}
\Phi[u] = \text{const.}(\{y_l\}_1^M) + <\phi_h (y-u),\phi_h (y-u)> + \tfrac{\alpha}{V M} <\rho_h u^{(n)},\rho_h u^{(n)}>
\end{equation}
To keep the notation simple, the definitions $\phi_h \equiv \phi^{1/2}$ and $\rho_h \equiv \rho^{1/2}$ are made.

  The solution can be found by setting the functional derivative to zero since the functional is positive and has a unique minimum as long as $\rho_h$ is bounded above zero everywhere in $\Omega$ and $\phi(t)$ contains at least $n$ mass points \cite{gkime71}.  We proceed as in Kimeldorf and Wahba \cite{gkime70} by using the scalar product invariance of the Fourier-Plancherel transform\cite{estei71} $\tilde g(\omega) \equiv F g = (2 \pi)^{-1/2} \int_\Omega {e^{i \omega t} g(t) dt}$, as well as the convolution theorem $F[a b] = (2 \pi)^{-1/2} \tilde a * \tilde b$ to get
\begin{equation}
\begin{split}
\Phi[\tilde u] = - 2 (2 \pi)^{-1/2} <F[\phi_h y],\tilde \phi_h * \tilde u>
+ (2 \pi)^{-1} <\tilde \phi_h * \tilde u, \tilde \phi_h * \tilde u> \\
+ (2 \pi)^{-1} \tfrac{\alpha}{V M} <\tilde \rho_h * ((-i \omega)^n \tilde u), \tilde \rho_h * ((-i \omega)^n \tilde u)>
+ \text{const.}
\end{split}
\end{equation}

  Functional differentiation with respect to $\tilde u(\omega)$ yields the minimum as the solution of
\begin{align}
\label{eq:FODE}
F[\phi y](\omega) &= F[\phi u] + \tfrac{\alpha}{V M} \omega^n F[\rho F^{-1}[\omega^n \tilde u]] \\
\label{eq:ODE}
\phi y &= \phi u + \tfrac{\alpha}{V M} (-1)^n \tfrac{\partial^n}{\partial t^n} (\rho u^{(n)})
.
\end{align}

  The last equation gives us the differential operator form of the minimization problem \eqref{eq:Phi} \cite{gkime71,gwahb90,fabra96}.  The interpretation for continuous sample density $\phi(t)$ is immediate, however it also remains valid for finite $M$ in the following sense.  On any interval between sample points, $\phi(t)$ as defined above is formally zero, and hence the solution $u(t)$ is that of the homogeneous equation -- i.e. a $2n-1$ order polynomial when $\rho$ is constant or (by definition) an L-spline for general $\rho$~\cite{gkime70}.  When integrating equation \eqref{eq:ODE} over small regions, we can treat all terms in the equation on equal footing at the sample points.  The net effect of the sample points is thus to effect changes in the function's $n^\text{th}$ (and higher) order derivatives.

  The above argument (and references) show that the unconstrained solution of Eq.~\eqref{eq:Phi} is a unique element of ${\mathbb H}^n(\Omega)$, and therefore (by definition) nonparametric.  To show precisely what is meant by a resolution independent spline estimate, we must prove that the spline function can approximate the nonparametric solution to arbitrary accuracy as the number of spline parameters increases.  To do this, we draw an analogy between the B-spline solution $\bar \theta | \alpha$ of Eq.~\eqref{eq:postth} and the Ritz-Galerkin finite element (weak) solution of Eq.~\eqref{eq:ODE}.  This lets us apply the typical error bounds for B-spline approximation \cite{ureif97}.

  Multiplying \eqref{eq:ODE} through by a test function $v(t)$ and integrating gives the bilinear form
\begin{equation}
\label{eq:bilinear}
a(u,v) = \int_\Omega \phi u v + \tfrac{\alpha}{V M} \rho u^{(n)} v^{(n)} dt
\end{equation}
%Where we have integrated by parts and identified the boundary conditions:
%\begin{equation}
%\left. (\tfrac{\partial^{n-k-1}}{\partial t^{n-k-1}} \{\rho u^{(n)}\} ) v^{(k)} \right|_{\partial \Omega} = 0, \quad k=0,\ldots,n-1
%\end{equation}
%which require continuity up to the $2n-1^\text{th}$ derivative for circular $\Omega$ and vanishing of the $n,\ldots,2n-1^\text{th}$ derivatives at $min\{t^l\},max\{t^l\}$ for infinite $\Omega$ to make a natural spline.

  To use standard procedures to get the approximation error in the induced norm $\|u\|_a = \sqrt {a(u,u)}$ we first define the norms
\begin{equation}
|v|_{k,m} \equiv \sqrt[m]{\int_\Omega {\left| \frac{d^k v}{d t^k} \right|}^m dt}
,
\end{equation}
with $|v|_{k,\infty}$ as the maximum of $v^{(k)}$ over $\Omega$.

  Now, let $u_h = \mathbf D(\{t\}_1^M) \cdot \bar\theta$ be the solution of Eq.~\eqref{eq:postth} and note that it is the minimizer of $\|u-u_h\|_a$ over all $u_h$ in the space of $r^\text{th}$ order Cardinal B-splines with knot spacing $h$ (denoted $\mathbb S^r_h$).  We thus have, for any $v_h \in \mathbb S^r_h$,
\begin{align}
\|u-u_h\|^2_a &\le \|u-v_h\|^2_a \notag \\
\|u-v_h\|^2_a &= \tfrac{1}{M} \sum_{l=1}^M (u(t_l)-v_h(t_l))^2
  + \tfrac{\alpha}{V M} \int_\Omega \rho (u^{(n)}-v_h^{(n)})^2 dt \notag \\
\|u-u_h\|^2_a &\le |u-v_h|^2_{0,\infty}
    + \tfrac{\alpha \rho_\text{max}}{V M} |u-v_h|^2_{n,2}
.
\end{align}
Where we have defined $\rho_\text{max} = |\rho_h (u-v_h)|_{0,\infty}$.
Choosing $v_h$ as a projection of $u$ onto $\mathbb S^r_h$ following Reif~\cite{ureif97}, we find
\begin{equation}
\label{eq:errest}
\|u-u_h\|^2_a \le (C_1 h^r)^2
  + \tfrac{\alpha \rho_\text{max}}{M} (C_2 h^{r-n})^2
,
\end{equation}
where $C_1 \text{ and } C_2$ are constants proportional to $|u|_{n,\infty}$.

  The approximation error thus falls into two regimes depending on the ``effective" samples per interval $h^n \sqrt{M/(\alpha \rho_\text{max})}$.  For large effective sample sizes, the first term in Eq.~\eqref{eq:errest} dominates and gives $\sqrt{\tfrac{1}{M} \sum_{l=1}^M (u(t_l)-u_h(t_l))^2} \sim \mathbb O(h^r)$. For small sample sizes, appropriate in the infinite resolution limit, the second term dominates and gives $|u-u_h|_{n,2} \sim \mathbb O(h^{r-n})$ -- both of which are of the same order as an optimal polynomial approximation to $u$.

\begin{figure}
%\showthe\columnwidth %convkerrn
\includegraphics[width=0.9\textwidth]{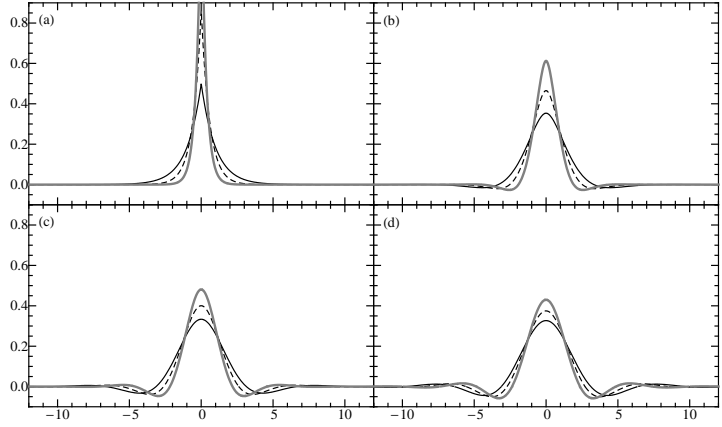}
\caption{Convolution kernels, $G(x-y)$, plotted for $\alpha/V M = \{1,1/3,1/9\}$ (solid black, dashed, thick grey, respectively), and $\phi=1$.  Higher sample numbers cause the kernel to steepen, placing more emphasis on $x=0$, while higher $\alpha$ values do the opposite.}
\label{fig:kern}
\end{figure}

  From Eq.~\eqref{eq:tG} it is also easy to find a convolution kernel estimate for $u$.  According to Eq.~\ref{eq:postth}, this estimate is
\begin{align}
\label{eq:conv}
u_h(t) &= M^{-1} \sum_{l=1}^M \func{G}{}{\frac{t-t_0}{h},\frac{t_l-t_0}{h}} Y_l \\
\label{eq:kern}
\func{G}{}{x,y} &\equiv B^r_c(x)^T \cdot \left(\tfrac{1}{M} \mathbf D^T \mathbf D + \tfrac{\alpha}{M} \mathbf Q\right)^{-1} \cdot B^r_c(y)
.
\end{align}
Which explicitly states our solution as a convolution with the kernel $\func{G}{}{x,y}$, so that our function estimate is a linear combination of B-spline basis functions.  Setting $\rho=1$ in the large sample size limit, when $\phi$ is approximately constant over a large range of knots, $\operatorname G$ is symmetric and its Fourier transform [from Eq.~\eqref{eq:FODE}] approaches \cite{bsilv84}
\begin{equation}
\label{eq:tG}
\tilde G(\omega) = \left(1+\frac{\alpha \omega^{2n}}{V M \phi} \right)^{-1}
.
\end{equation}
Convolution kernels for other several derivative orders, $n$, are plotted in Figure~\ref{fig:kern}, while for general $\rho$ and $\phi$, Abramovich and Grinshtein~\cite{fabra99} have provided a method to systematically derive such asymptotic approximations.

  These asymptotic approximations give a sense of how the algorithm changes in response to differing penalty orders, $n$.  From the discussion following \eqref{eq:ODE}, as $n$ increases, higher order derivatives of the function are changing at the sample points, leading to a larger width convolution kernel.  Note that the asymptotic approximation plotted in Fig.~\ref{fig:kern} breaks down when $M<p$.  However resolution independence is still maintained; since in intervals without design points the solution will approximate the L-spline solution to \eqref{eq:ODE}.  This provides a second interpretation to $n$ as specifying the order of the ``default" polynomial solution.

%	Asymptotic limits of the MSE can be found using familiar properties of Toeplitz matrices and the Fourier transforms found above.  The first term in \eqref{eq:MSE} is bounded by
% TODO: missing V in some equations here...
%\begin{equation}
%\left\|\mathbf D \cdot \mathbf K \cdot \tfrac{\lambda}{z M} \mathbf Q \cdot \theta_0 \right\|^2
% \le |\theta_0|^2 \max_{\omega \in [0,2 \pi]}{\frac{\phi M \tilde B^2 (\alpha \omega^{2n}/M)^2}{(\phi + \alpha \omega^{2n}/M)^2}}
%\end{equation}
%Which is monotonically increasing over its range in $\omega$, having its maximum at $2 \pi$.  The total contribution from this term is thus $\mathbb O(\frac{\alpha^2 |\theta_0|^2}{\phi M^2})$.  The second term can be approximated using Szego's theorem as
%\begin{equation}
%\mathrm{Tr}\left[ (M^{-1} \mathbf D \cdot \mathbf K \cdot {\mathbf D}^T )^2 \right]
% \rightarrow \frac{p}{2 \pi} \int_0^{2 \pi} (1 + \frac{\alpha \omega^{2n}}{\phi M})^{-2} d\omega \approx \mathbb O(p)
%\end{equation}
%Thus the total contribution from the second term is $\mathbb O(p/M)$ and dominates the asymptotic MSE.  The apparent resolution-dependence of this term is due to the implicit assumption in Eq.~\ref{eq:tDTD} that the design matrix is full-rank.  Our numerical calculations show that the observed MSE is not resolution-dependent.

\subsection{ Marginal distribution of the variance}
\label{sec:zmargin}

  As in the previous section, the change of variables from $\lambda$ to $z \alpha$ uses the insight that the spline parameter estimate, $\bar \theta$, depends only on the ratio $\alpha$.  This change makes the posterior PDF of section~\ref{sec:post} into
\begin{equation}
\begin{split}
\func{P}{}{\theta z \alpha | D I} &\propto \alpha^{(p-n)/2-1} z^{(M+p-n)/2-1} \\
  &\times \exp -\tfrac{z}{2}\left\{ 
      \left\|Y-\mathbf D \cdot \theta\right\|^2 + V_0
      + \alpha \left( \theta^T \cdot \mathbf Q \cdot \theta + E_0 \right)
  \right\}
.
\end{split}
\end{equation}

  The marginal probability density $z \alpha | D I$ is found by integrating over the $p$-dimensional $\theta$, assuming there are enough input measurements to determine the $n$ improper dimensions of $Q$ corresponding to an $n-1$ degree polynomial fit.
\begin{align}
\func{P}{}{z \alpha | D I} &= \int \func{P}{}{\theta z \alpha | D I} d\theta \notag \\
\label{eq:zamarg}
 &\propto
  \frac{\alpha^{(p-n)/2-1} z^{(M-n)/2-1}
        \efunc{-\tfrac{z}{2}\left(\epsilon^2_f + V_0 + \alpha (\epsilon^2_Q+E_0) \right)}}
  {\abs{{\mathbf D}^T \cdot \mathbf D + \alpha \mathbf Q}^{1/2}} \\
 &\epsilon^2_f \equiv \left\| \mathbf D \cdot \bar \theta - Y\right\|^2,
  \epsilon^2_Q \equiv {\bar\theta}^T \cdot \mathbf Q \cdot \bar\theta \notag
\end{align}

  This shows that the variance is Gamma distributed conditional on $\alpha$.
\begin{equation}
\label{eq:z|a}
\func{P}{}{z | \alpha D I} \sim \Gamma(\frac{M-n}{2}, \frac{\epsilon^2_f + V_0 + \alpha (\epsilon^2_Q+E_0)}{2})
\end{equation}
  Interestingly, this estimate for the variance is at odds with most procedures suggested in the literature \cite{gwahb90}.  Taking the inverse of the average, we get $\func{E}{}{z|\alpha}^{-1} = [\epsilon^2_f + V_0 + \alpha (\epsilon^2_Q+E_0)]/(M-n)$, which resembles the square error estimate with $M-n$ samples but includes the string energy, $\epsilon^2_Q$.  An explanation for this could be that specifying $\alpha$ gives information about the error scale, $z$ relative to average string deviation where specifying $\lambda$ alone would not.  This reveals some of the subtle distinctions which are often missed between $\lambda$ and $\alpha$, especially in discussions of variance estimates.  It also validates the interpretation of $E_0$ and $V_0$ as residual uncertainties and proves that the variance estimate is insensitive to them as long as they remain less than $\epsilon^2_Q$ and $\epsilon^2_f$.

%  We can use the above marginal PDF to implement a more efficient MCMC sampling method via drawing from the joint conditional distribution $\func{P}{}{z \theta | \alpha D I} = \func{P}{}{z | \alpha D I} \func{P}{}{\theta | z \alpha D I}$ followed by sampling $\alpha | \theta z D I$.  Sampled statistics can be greatly improved by collecting the conditional expectation and variance of $\theta | z \alpha D I$, $\alpha | \theta z D I$, and $z | \alpha D I$ at each iteration of the sampler.  These collected values can be used to numerically approximate the full posterior for $\theta$ during subsequent functional inference.
%However,  since $\alpha$ cannot be so easily defined for the vector-valued $f(r)$ regression problems considered later, conditional average and variance statistics were only collected for $\theta$ in the calculations done for this report.

\subsection{ Marginal distribution of the Penalty Parameter}
\label{sec:amarg}

  Since $z | \alpha D I$ has a Gamma distribution, the normalizing constant for $z$ is known and it can be integrated out of \eqref{eq:zamarg} to give
\begin{equation}
\label{eq:amarg}
\func{P}{}{\alpha | D I} \propto \frac
  {\alpha^{(p-n)/2-1} \left(\epsilon^2_f + V_0 + \alpha (\epsilon^2_Q+E_0)\right)^{-(M-n)/2}}
  {\abs{{\mathbf D}^T \cdot \mathbf D + \alpha \mathbf Q}^{1/2}}
.
\end{equation}

  The first term and the determinant in the denominator can be thought of as offsetting terms, pushing smoothness higher until the point where the $p-n$ eigenvalues of $\mathbf Q$ dominate the determinant.
%In the uniform sampling density and $\rho$ case, the hat matrix is a T{\"o}plitz matrix for an infinite measurement domain or a circulant matrix for a periodic one.  This analogy lets us use Sz{\"e}go's theorem along with \eqref{eq:tG} to get $\ln \abs{{\mathbf D}^T \cdot \mathbf D + \alpha \mathbf Q} \approx \tfrac{p}{2 \pi} \int_0^{2 \pi}{\ln (\phi + \alpha \omega^{2 n}) d\omega} \approx p \ln (\phi + \alpha (2 \pi)^{2 n})$. % TODO - ck. last approx. against lower bound

  The second term is the most interesting if we remember that $\alpha$ represents the ratio between the spline tension and spring constants.  The above distribution says that we can learn about the over/under-smoothing tradeoff through comparing the relative magnitudes of the function deviations, $\epsilon^2_f(\alpha)+V_0$ and the roughness, $\epsilon^2_Q(\alpha)+E_0$.  As $\alpha$ increases, the function deviations increase and the roughness decreases, so that the two energy scales can reach a balance.  The appearance of $E_0$ and $V_0$ is natural in this context, since it prevents choosing either extreme.

  The (twice negative log of the) posterior likelihood~\eqref{eq:amarg} can also be compared to the AIC and GCV functions for choosing $\alpha$ which have been extensively analyzed in the literature.
Defining $\mathbf H(\alpha) \equiv \mathbf D \cdot (\mathbf D^T \cdot \mathbf D + \alpha \mathbf Q) \mathbf D^T$, these two functions are:
\begin{align}
\label{eq:GCV}
GCV(\alpha) &= M \epsilon^2_f \left( M-\mathrm{Tr}(\mathbf H) \right)^{-2} \\
\label{eq:AIC}
{AIC}(\alpha | \hat z) &= \hat z \epsilon^2_f + 2 \mathrm{Tr}(\mathbf H) 
.
\end{align}
Where we have set $\hat z = 23.74^{-2}$, following Eilers and Marx~\cite{peile96} and ignored constant terms.

  First, we note that neither of these two functions has a straightforward generalization for inference on vector-valued functions, so we must consider alternatives here.  This is because there is not a one-to-one correspondence between individual functions to be added together and the dimensions of $y(r)$, rather all functions have the ability to contribute to all output dimensions.  

  Examining the criteria specified in this report, the GCV has very good properties for the one-dimensional case.  Although the interpretation of $\alpha$ as specifying the relative scale between $\lambda$ and $z$ is not clear with the GCV estimate, its minimum remains invariant to scaling both the function and its error (scale independence) since $\epsilon^2_f$ and $\mathbf H$ are functions of $\alpha$ only.  However, alternatives such as the AIC or Mallow's $C_p$ criterion which require an {\it a priori} estimate for the variance do not necessarily have this property.

\begin{figure}
%\showthe\columnwidth %helmet
\includegraphics[angle=-90,width=0.9\textwidth]{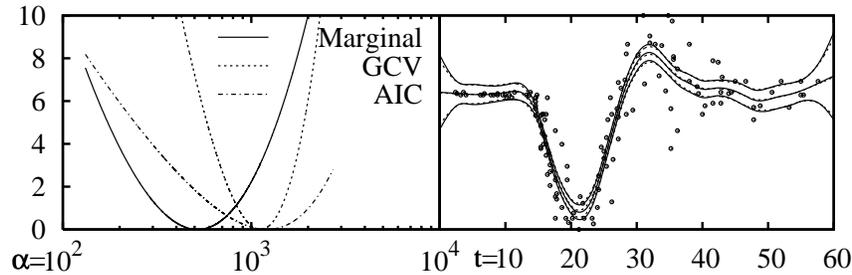}
\caption{ Comparison of alternate criteria for estimating $\alpha$.  In the left panel, GCV and AIC are as in Eq.s~\eqref{eq:GCV} and~\eqref{eq:AIC}, while marginal denotes $-2 \log P(\alpha | D I)$, and all functions have been translated to set the minimum to zero. The right panel shows the posterior average and uncertainty using the present method and GCV.}
\label{fig:helmet}
\end{figure}

  Figure~\ref{fig:helmet} compares the different functions of $\alpha$ for H{\"a}rdle's motorcycle helmet data~\cite{peile96}.  It can be seen that the GCV and AIC predict values in roughly the same neighborhood, while the present log-likelihood has a minimum at a much lower $\alpha$.  Despite this, the present method differs very little from the GCV estimate in this example, while the AIC estimate (not shown) is indistinguishable from the GCV curves.  We can therefore conclude that averaging over a range of $\alpha$ values does not spoil the fit to this data set, but allows additional smoothing (as can be seen from Fig.~\ref{fig:kern}) as well as a more cautious error estimate.

\section{ Example Problems}
\label{sec:probs}
  We consider some example cases which serve to test the scale independence of our method as well as illustrate the significance and usefulness of multidimensional function matching.  First, the proposed prior is compared to other suggestions in the literature for a few simple test functions.  Next, we progress to a high-dimensional example which attempts to describe the dynamics of 256 atoms.  This system and others like it are numerically tractable since their behavior can be explained in terms of only a few additive functions, while the main difficulty of fitting these functions lies in separation of the measured sums.

\subsection{ Method Comparison}

  We investigated the properties of various P-spline formulations using a standard set of model functions proposed by Lang and Brezger~\cite{slang04}.
As in that report, $\rho$ is chosen to be constant, and the derivative order is set to $n=2$.  The most important test functions for our current purposes are the linear function $f_1(r) = r/1.758$ and the sinusoidal function (here modified from the original for exact periodicity) $f_3(r) = \sin(\pi x/3)/0.72$.
The linear function tests the algorithm's stability in the degenerate polynomial solution case, and the sinusoidal function tests the algorithm's over/undersmoothing trade-off.
The sinusoid function was modified because for the small number of samples used here almost all fits were noticeably linear when periodic B-splines were not used.
These functions are defined over the range $[-3,3]$ and fit to 20 4$^{\text{th}}$ order (cubic) B-spline knots using 20 equidistantly spaced noisy samples in order to force the solver into the low-sample regime.  

  In order to compare across different versions of the algorithm and different scales within each version, the same set of independent random noise values generated from the standard normal distribution were appropriately scaled and used in all tests.  All mean squared error results in this subsection have been renormalized via division by the square of the initial scaling applied.  This makes for easier comparison, but requires us to keep in mind that larger error scales still produce worse fits in an absolute sense.  All systems used MCMC Gibbs sampling of $\lambda$,$z$,$\theta$ as described in section~\ref{sec:post} with a burn-in of 2500 steps, followed by 25000 sampling steps, collecting one sample every 5 steps for a total of 5000.  Correlation times for $\lambda$ were always higher than those for $z$, and are included in our figures (right y-axis).  Correlation times for $\theta$ are not as important, since the conditional average ($\func{E}{}{\theta|\lambda,z}$) has been collected with each sample.

\begin{figure}
%\showthe\columnwidth % 1
\includegraphics[width=0.4\textwidth]{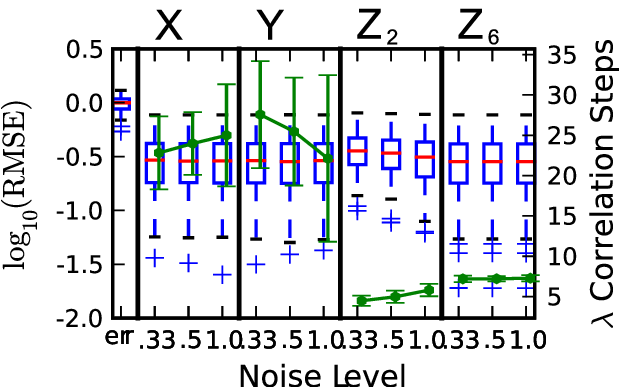}
\includegraphics[width=0.4\textwidth]{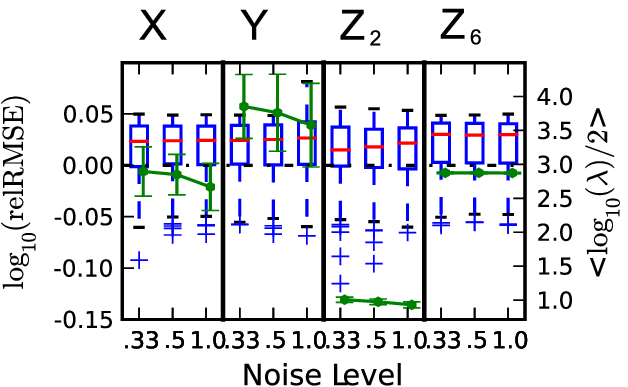}
\caption{Boxplots of error distributions for (left to right) variance of input samples, $X$: this report's method, $Y$: Lambert and Jullion's method, and Lang and Brezger's method using $Z_2$: $b=10^{-2}$ and $Z_6$: $b=10^{-6}$.  For each method, several noise scales are shown as explained in the text.  Green lines (offset for visual clarity) display decorrelation time.  The right panel shows the ratio between the sample RMSE and the estimated RMSE (left scale, calculated by averaging $z^{-1}$) and the magnitude of posterior average penalty parameter.}
\label{fig:1}
\end{figure}

\begin{figure}
%\showthe\columnwidth % 3
\includegraphics[width=0.4\textwidth]{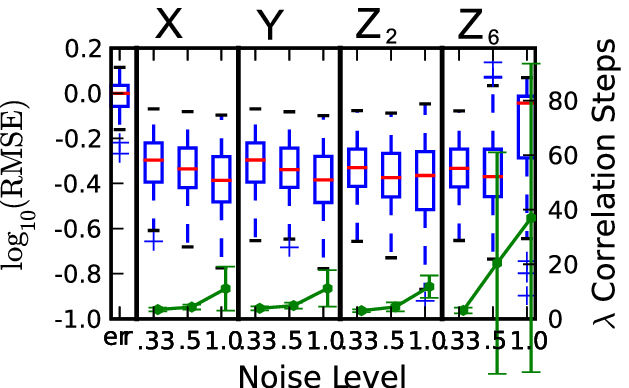}
\includegraphics[width=0.4\textwidth]{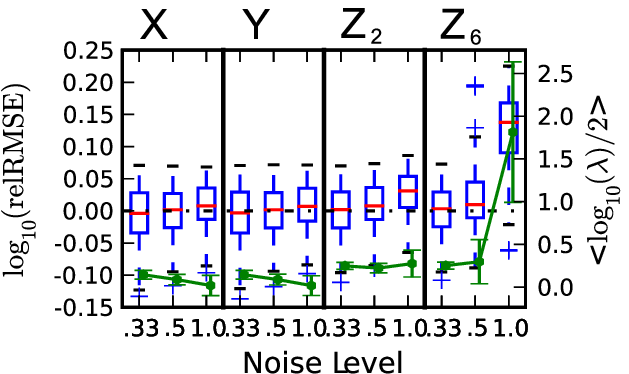}
\caption{Boxplots of error distributions, labeled as in Fig.~\ref{fig:1}.}
\label{fig:3}
\end{figure}

  Figures~\ref{fig:1} and \ref{fig:3} show boxplots of the (normalized) empirical root mean squared deviation (RMSE) of the spline from its respective function at the 20 input sample points -- log scale, left axis.  Each figure shows the RMSE for the input samples (far left) as well as results from four different prior distributions for $\lambda$.  They are (from left to right): $X$: $(\lambda | I) \sim \Gamma(0,10^{-10}/2), (z | I) \sim \Gamma(0,10^{-10}/2)$; $Y$: $(\lambda | \delta I) \sim \Gamma(1,\delta), (\delta | I) \sim \Gamma(10^{-4},10^{-4})$; $Z_2$: $(\lambda | I) \sim \Gamma(1,10^{-2})$, and $Z_6$: $(\lambda | I) \sim \Gamma(1,10^{-6})$.  For each function and prior distribution, several different noise scales were applied to the random deviations, with $\sigma$ indicated on the x-axis.  In order to compare the relative error achieved with the efficiency of the method, the autocorrelation function of $\lambda$ was calculated and fit to an exponential for each test to estimate the correlation time.  The average and standard deviation of this quantity over all 100 runs gives an indication of the number of MCMC steps required to draw an independent sample.

  The linear case is an interesting smoothing limit to investigate, since the best approximation we can make is fitting to a polynomial of order $n-1$, and the bias of the MSE derived in Eq.~\eqref{eq:MSE} is zero.  A polynomial fit corresponds to $\lambda \rightarrow \infty$ in Eq.~\eqref{eq:sform}, a case for which our numerical solver becomes unstable [see discussion in section~\ref{sec:pprior}], which indeed occurred during our test when we set $E_0=0$.  Our choice for $E_0$ sets an effective maximum on $\lambda$, which could lead to undersmoothing.  However, the MSE and RMSE results of Fig.~\ref{fig:1} from different methods are almost identical with the exception of $Z_2$, indicating that $\lambda$ was large enough.  This figure shows that the $\lambda$ selected by each method is strongly influenced by their respective prior distribution.  In particular, $Z_2$ and $Z_6$ show a clear preference for $10^{-2}$ and $10^{-6}$, respectively.  Methods $X$ and $Y$ seem to show more response to variations in the error scale, decreasing $\lambda$ as $\sigma^2$ increases (implying a relatively stable choice for $\alpha$).

  Fitting to the sinusoid function shows the under/oversmoothing tradeoff of each method.  Figure~\ref{fig:3} shows that all methods (with the exception of $Z_2$ and $Z_6$) perform equally well predicting both the function and error scale.  Methods $Z_2$ and $Z_6$ break down at low signal to noise, where the strength of their prior distribution for $\lambda$ begins to outweigh the data.  The similar performance of $X$ and $Y$ in this test indicates that they are less sensitive to their prior for $\lambda$.  This could have been predicted from the fitting results, since (for $X$) the function error, $\epsilon^2_f$, deviates by more than $E_0$ from a polynomial solution (see Sec.~\ref{sec:amarg}), and (for $Y$) $\lambda$ does not have too small a value.

  To see why the fitting results of $Y$ may display sensitivity to the prior parameters, we can integrate the compound prior $\int \func{P}{Y}{\lambda \delta|I} d\delta$ to give $\func{P}{Y}{\lambda|I} = (b/(\lambda+b))^a$ with $a=b=10^{-4}$ as used here.  This function deviates significantly from the Jeffrey's prior $\lambda^{-1}$ at small $\lambda$, indicating a preference for larger $\lambda$.  However, it has a smoother decay toward zero at $\lambda \rightarrow \infty$ than $X$.  This smoother decay at large $\lambda$ could also be achieved by adding a similar hyperprior for $E_0$ in our method.

  The function $f_3$ was selected for a further test of scale-dependence.  This was done by varying the overall scale of the function {\em and} its noise on a logarithmic scale from $10^{-9}$ to $10^{+9}$.  Each data set used for fitting contained 50 equally spaced samples.  Although a scale-dependence of the spline fit was noted by Lang and Brezger with the suggested remedy of standardizing the input samples, this strategy is difficult to justify if adaptive penalties are considered, and impossible for multi-dimensional measurements, where no linearly independent set of directions to standardize is guaranteed.  None of the tests reported here make use of such standardization of input values, since an ideal method should produce identical (scaled) results for any choice of overall scale.

\begin{figure}
%\showthe\columnwidth % S
\includegraphics[width=0.4\textwidth]{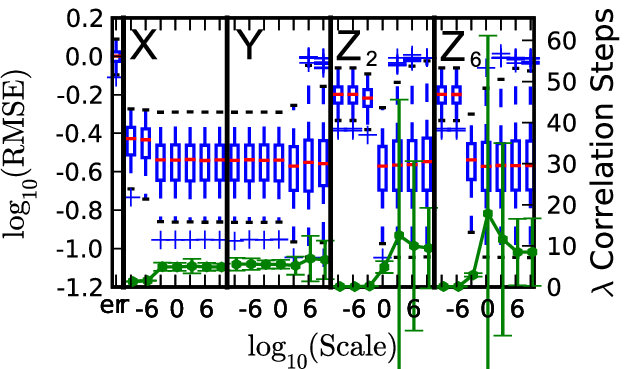}
\includegraphics[width=0.4\textwidth]{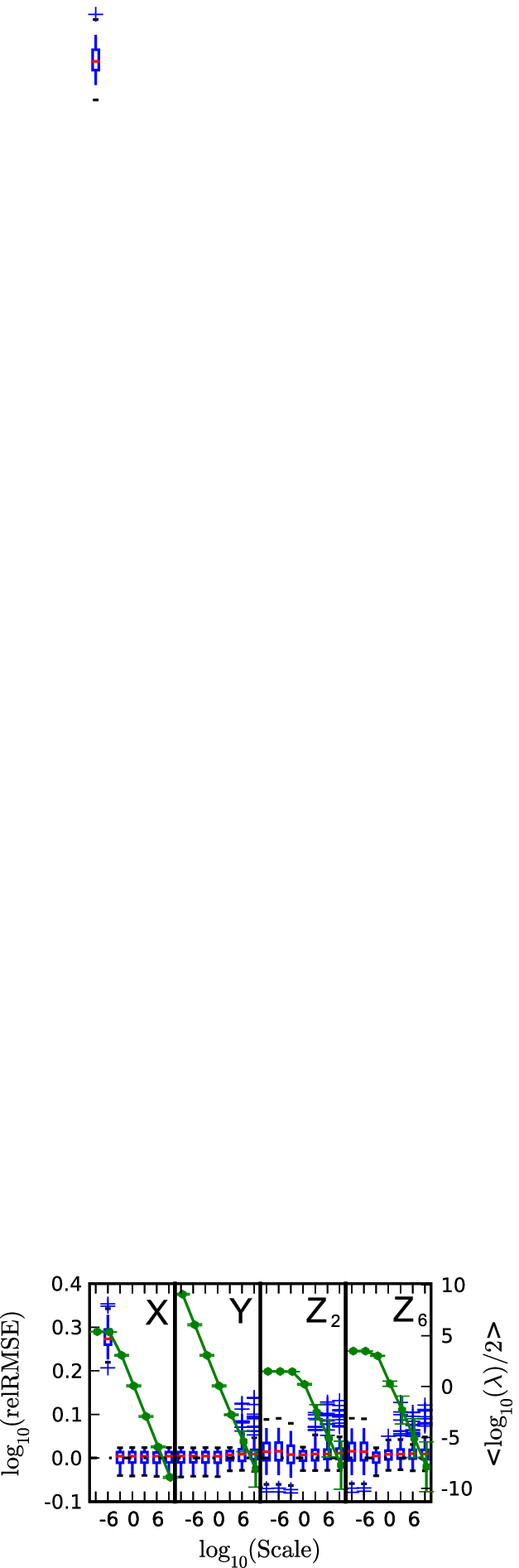}
\caption{Boxplots of error distributions showing variation with respect to total problem scale ($10^{-9}$ to $10^9$), other symbols as in Fig.~\ref{fig:1}}
\label{fig:S}
\end{figure}

  The results of this test are shown in Fig.~\ref{fig:S}.  As expected, $Z_2$ and $Z_6$ show a bias toward their respective prior $\lambda$ values, giving good MSE results for scales of $10^{-1}$ or $10^{-3}$.  However, they show under/oversmoothing when the scale is below/above that value, respectively.  Undersmoothing has a characteristically higher MSE and a larger range of $\sigma$ estimates, while oversmoothing shows a wider range of MSEs and outliers where $\sigma$ is overestimated along with a wide range of $\lambda$ correlation times.

  For $X$, the method breaks down at scales less than $10^{-5}$.  This happens because the scale of the input data is below the order of $\sqrt {E_0}$, violating our assumption in setting $E_0$.  To understand this in the context of our prior from Sec.~\ref{sec:pprior}, even as the data scale diminishes (reducing $\epsilon^2_Q$), $\lambda$ gets stuck at $E_0$ and can't go higher.  This can be seen directly in Fig.~\ref{fig:S}.  The same process happens for $z$, which gets stuck at $V_0$ as the scale diminishes.  This has caused $\log_{10} \hat \sigma^2$ to bottom out at $-11.5$ at a scale of $10^{-6}$ and $-11.6$ at a scale of $10^{-9}$ [off the scale of Fig.~\ref{fig:S}].  Based on these observations, we can ask if $\epsilon^2_Q/E_0$ or $\epsilon^2_f/V_0$ is too small (say less than $10$) as an indicator of whether our method is predicting an exact polynomial fit or an exact fit of the input data.  At large scales, these ratios are large, indicating that $E_0$ and $V_0$ have become irrelevant and scale-independence is achieved.  As explained above for method $Y$, $\lambda$ must decrease at large scales, but the prior for $\lambda$ is too small close to zero, causing oversmoothing.

\subsection{ Multidimensional Problems}

  Modeling of dynamic processes is usually carried out via numerical integration in time ($\tau$) of, e.g. Newton's equation of motion for a given potential function, $E(r)$.
\begin{equation}
\label{eq:newton}
m \frac{\partial^2 r_i}{\partial \tau^2} = -\frac{\partial \func{E}{}{r}}{\partial r_i} \equiv f_i(r)
\end{equation}
Where the system position is specified by $r \in \mathrm{R}^N$ ($N$ being three times the number of atoms, $r$ being indexed by $3i+\alpha$).  Standard molecular dynamics potential functions consist of multiple additive functions and are of the form
\begin{equation}
\func{E}{}{r} = \sum_{k=1}^{N_f} E_k(t_k(r))
.
\end{equation}
where each $t_k(r)$ is a scalar function of $r$, e.g. a distance between two atoms.
Attempting to match samples of the force $\{y,r\}$ requires differentiating the above with respect to $r$ to give the linear mixed model
\begin{equation}
\label{eq:ndsamp}
f(r) = \sum_k -\frac{\partial t_k(r)}{\partial r} {E'}_k(t_k(r)) + \mathrm{noise}
.
\end{equation}
Here $-\partial t_k(r) / \partial r$ is an $N$-dimensional vector with components $-\partial t_k(r) / \partial r_{3i+\alpha}$ giving the direction of each contribution to the total force, identified with $g_k(r)$ in section~\ref{sec:multidim}, and each $E'_k(t_k(r)) = \mathbf B_k \cdot \theta_k$ is modeled via a B-spline basis.  The most striking property of this model is that it holds to the simple linear mixed model form and is amendable to the same sorts of analyses carried out there.

  For a nontrivial example of this type of problem, we consider a system of $256$ Lennard-Jones particles with unit mass interacting via the pair potential
\begin{equation}
\label{eq:PE}
\func{E}{}{r} = \sum_{1 \leq i<j \leq 256}
  4 c_{ij} \left( |r_i-r_j|^{-12} - |r_i-r_j|^{-6} \right)
.
\end{equation}
Using the notation $|r_i-r_j|$ to mean the Euclidean distance between $r_i \in \mathrm R^3$ and the closest periodic image of $r_j$ (since all the atoms have been placed into a cubic cell of length $7.49$).  By setting $c_{ij} = 1/2$ whenever $1 \le i \le 128 \text{ and } 129 \le j \le 256$, or $c_{ij} = 1$ otherwise, we have set up a two-component (binary) mixture.  Due to their relatively low cross-attraction, particles of type A ($1 \le i \le 128$) and those of type B ($129 \le i \le 256$) have been found to be immiscible under the types of conditions used here~\cite{kmaed03}, separating into two liquid layers within the simulation cell.  The liquid state may be metastable with respect to a solid crystal phase, but such crystallization was not observed in any of the simulations reported here.

  Numerical integration of Eq.~\eqref{eq:newton} with a time-step of $1.461 \times 10^{-3}$ was used to simulate this system until steady-state behavior was observed.  Five hundred configurations, $\{r\}_1^{500}$, were obtained by sampling every $500$ steps in a canonical ensemble simulation, which uses a slight (stochastic) modification of the equations of motion~\cite{rskee02} to guarantee that the velocities $\partial r_i / \partial \tau$ are independently normally distributed about zero with variance $\beta^{-1}$, here chosen to be $0.7917$.  Although the units stated in this report make use of a reduced units convention, the problem remains unchanged if we set physical units of $\epsilon=1.26$ kJ/mol, $\sigma=3.7$ \AA, $T=120$ K, $m=16$ g/mol, $\Delta \tau = 1.935$ fs, and $\gamma = 5.1677$ ps$^{-1}$.

  For consistency with Eq.~\eqref{eq:ndsamp}, the set of forces was generated from the set of configurations using the following procedure.  First, the three pairwise potential functions $4 c_{ij} \left( t^{-12} - t^{-6} \right)$ were replaced with their spline representations on $t \in (4/7,17/7)$ and used to find the mean force corresponding to each configuration.  Next, independent normally distributed random noise with magnitude $\sigma_F = 60.91$ was added to each dimension of each generated force sample.  This procedure generates the same sample distribution as would be expected from a Langevin dynamics simulation with a moderate damping coefficient of $\gamma \Delta \tau = 10^{-2}$.

  Since we have three functions to match (A:A, A:B, and B:B), we use Eq.~\eqref{eq:multfn} with $N_F=3$ to compute $\mathbf D_{l,k} \in \mathbb M_{N \times p_k}$ for each frame and calculate the posterior mean and covariance using Eq.~\eqref{eq:postthm}.  In this equation, the sets $K$ are assumed to include all terms of Eq.~\eqref{eq:PE} which share a common set of parameters $\theta_K$ (i.e. interactions between atoms of type A:A, A:B or B:B).  Similarly, we will assume each particle type has its own $z_i$, so we use Eq.~\eqref{eq:postzm} with $N_I=2$ and each set $I$ is simply the set of all coordinates belonging to atoms a common type.
\begin{figure}
%\showthe\columnwidth % sample
\includegraphics[angle=-90,width=0.9\textwidth]{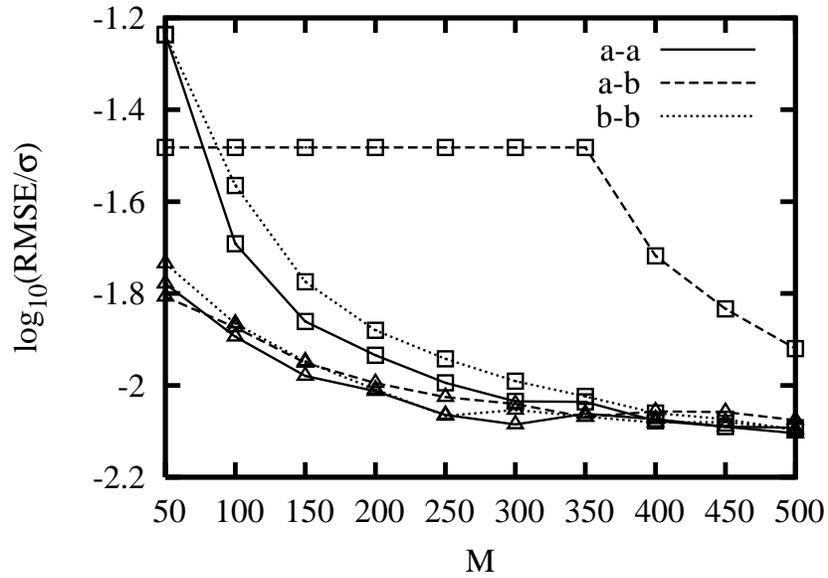}
\caption{Effect of sample size on average error of the fitted functions.}
\label{fig:sample}
\end{figure}
\begin{figure}
%\showthe\columnwidth % his
\includegraphics[angle=-90,width=0.9\textwidth]{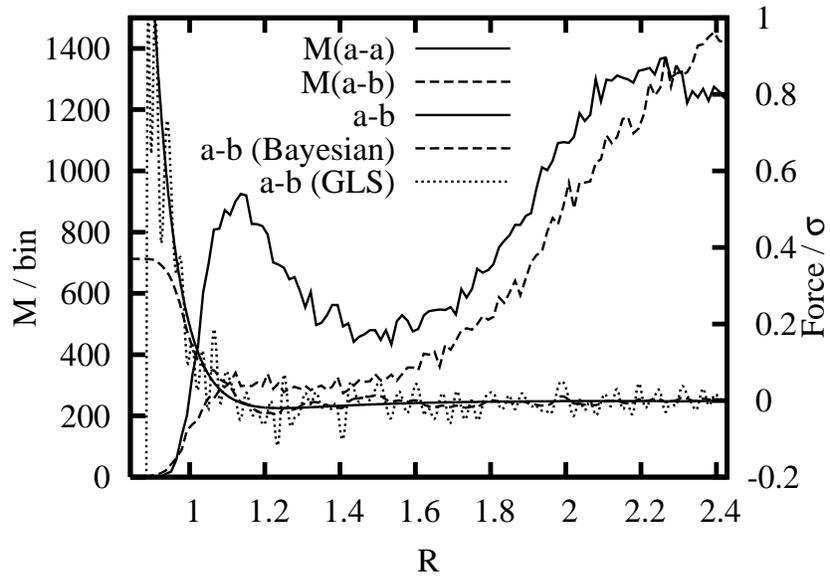}
\caption{Effect of sample size on distribution of observed distances.}
\label{fig:his}
\end{figure}

  The pairwise functions were fit to $6^\text{th}$ order B-splines with $170$ knots and $h=0.1/7$ to give a range of $(0,17/7)$, forcing the function and all its derivatives to zero at $17/7$.  MCMC sampling was carried out as described for the one-dimensional test cases except for the use of the density function $\rho(r)=r^2$, which is more appropriate for the spherically symmetric function domains considered here.  Figure \ref{fig:sample} shows the fitting results and a comparison between the posterior average and maximum likelihood estimators for a variety of sample sizes.  Corresponding distributions of the observed pairwise distances for the $M=50$ case are shown in Fig.~\ref{fig:his} as a function of sample size (left scale).  For this case the present approach is contrasted with the generalized least squares solution on the right scale.

  Since the actual system does not allow observations of all pairwise distances (particularly for small $r$), the average mean-squared error between the functions and their spline fits were calculated by averaging over the observed distances from all $500$ samples.  The range of observed distances is also the appropriate one for considering the approximation error estimates of section~\ref{sec:res}, which serve as upper bounds on any subset of $\Omega$ where $\rho>0$.

  An unusually high error is observed for the function describing interactions between particles of type A and B.  Inspection of the spline estimates reveals that this error is due to oversmoothing (the MLE estimate was essentially zero) caused by the relatively small number of samples for this function, and so does not occur when the prior is set to zero -- even for the $M=50$ case (GLS, Fig.~\ref{fig:his}).  This failure of the MLE makes it a worse choice than GLS, and could not have been predicted from calculation of the posterior function error.  The expected error conditional on the MLE estimate for $\alpha,z$ is (in units of Fig.~\ref{fig:sample}) $-5.0$ for all functions at $M=50$.  At $M=100$, the A-A and B-B error estimates jump to around $-1.9$ and remain constant as $M$ increases, while the A-B error estimate remains at $-5.0$ until $M=350$, where it jumps to $-2.2$ and slowly increases to $-2.1$ at $M=500$.  These numbers significantly underestimate the fitting error at small sample sizes due to their neglect of variations in $\alpha,z$.

  On the other hand, the posterior average estimate behaves as expected, approximating the input function with accuracy increasing with sample size.  The expected function error using this method is slightly over-estimated due to the uncertainty in $\alpha,z$ -- smoothly decreasing from $-1.5,-1.6$ for A-A,A-B at $M=50$ to $-1.9,-1.8$ at $M=500$.  Examining the $M=250$ case, the likelihood ratio of the posterior average estimate $\bar \theta$ to the MLE is $10^{-585}$, but the average estimate performs better, and must be used, because of the width of the posterior distribution.  Considering $\theta$ as a $510$-dimensional vector, if the probability distribution for $\alpha | \bar \theta$ is relatively flat over a range $R$ away from $\bar \theta$, then there are $R^{510}$ ``states," $\theta,\alpha,z$, for which $\theta = \bar \theta | \alpha$ and which therefore have a very low, but similar posterior probability.  Integrating over a region in state-space is essential in justifying the astronomical difference in pointwise probabilities.  Finally, as the sample size increases, the posterior probability narrows, causing both estimators to converge.  The dramatic failure of the MLE shown in Fig.~\ref{fig:sample} demonstrates the importance of averaging over the posterior parameter distribution for small sample sizes.

\section{ Discussion}
\label{sec:disc}

  This paper presents a re-derivation of the commonly employed Bayesian statistical model for penalized spline matching.  Several questions are addressed relating to the applicability of the method for general problems.  Its main limitations stem from the possibility of choosing an inadequate form for the function to be fitted, $y(r)$.  This can happen either by leaving out the dependence of $y(r)$ on some important function of $r$ -- causing larger than necessary noise in the data set -- or by choosing a resolution that is too low (too large bin width) in order to save computational time.  This latter problem can result in an oversmoothed function and will again increase the fitted value of $\sigma^2$.

  Physical concepts are important to justify use of the quadratic penalty parameter and give a useful interpretation to the fitting process.  Using the energy function for states of a stretched string, we can apply the Boltzmann distribution to derive a simple prior probability on function space.  This method can be used to justify further adaptations of Eq.~\eqref{eq:stren} and could prove to be generally useful in other problems of inference.  Also by analogy to the physical system, we can anticipate a unique solution to the optimization problem Eq.~\eqref{eq:phi}.  This proof is a classical result of the smoothing spline literature, and has been invoked here to show that our B-spline representation also optimally approximates the unique minimum, converging as the resolution is increased.
%  Another arbitrary parameter, $n$, is used for imposing a chosen degree of smoothness on the spline estimate and can be seen as either widening the effective sample bin width or specifying a default shape for the solution in the absence of data.

  By explicitly considering the variation of the smoothing parameter with respect to function scale, we are able to clearly present and compare several choices for the smoothing parameter proposed in the literature.  These classical choices are difficult to apply in the multidimensional context, and several have the additional possibility of scale dependence.  Although the GCV criterion has the desired scale independence property, it also has the problem of predicting an exact match to the input data at low sample sizes \cite{gwahb95}.  This problem (and the related one of predicting an exact polynomial fit) was solved in this report by placing constraints on the ability of the smoothing parameter estimate to force such an exact fit.  The result is an estimator which is asymptotically scale independent for input data satisfying these minimum variance and polynomial deviation properties.

  Special consideration has also been given to formulation of vector-valued function fitting problems.  An important observation is that the dimensionality of these problems can make maximum likelihood estimation ineffective for small sample sizes.  There have certainly been previous treatments of multidimensional fitting and integrator matching, both in considering vector generalized additive models~\cite{tyee96} and in adapting generalized linear least squares to use Bayes' theorem in physical systems~\cite{ghumm05,pliu08}.  However to the author's knowledge, this is the first report of the extension of P-splines to fit the drift and diffusion terms of molecular Langevin dynamics simulations.

  The approach developed in this paper can be directly used to fit stochastic integration processes such as the Langevin systems considered here, giving a well-defined way to extrapolate forward in time.  This would work particularly well since the inference result gives the average and variance of each step in the integration process.  However, although normally distributed force error is usually assumed in applications of the Langevin equation (and Euler integration of stochastic equations in general), the question of correspondence between dynamics where this assumption does not hold requires separate consideration.

  Future research could also re-consider the creation of an ``overall" scale-invariant prior for smooth $z(r)$ where random noise is an issue in this process.  This would be particularly important for variable volatility in market data or inhomogeneous molecular systems.

\appendix

\section{ Integrated Spline Derivatives}

  We present a method to calculate the integral \eqref{eq:Q} using B-spline basis functions~\cite{cboor78} for general $\rho$ and $n$.  Klaus H{\"o}llig has presented a simpler method for this calculation when $\rho$ is constant in Ref.~\cite{kholl03}.  For general $\rho$ we proceed by expressing the $r^{\text{th}}$ order B-spline $B(x;\theta)$ as a set of piecewise polynomials and calculate the contribution to the integral from each interval separately.

  To begin, we re-scale the spline value $x$ to $u=\tfrac{x-x_0}{h} + \Delta$ which maps the spline range $x\in(x_0,x_0+L)$ to the interval $u\in(\Delta,L/h+\Delta)$.  Here, $\Delta$ can be either $r/2$ for a periodic spline or $r-1$ for aperiodic splines, where setting $L/h$ past $p-\Delta$ forces successive derivatives of the spline to go to zero at $L$.  Using this mapping, spline knots are placed at $[0,1,\ldots,p-1]$ to give the representation in terms of the B-spline basis function $M_r(u)$.
\begin{equation}
B(u;\theta) = \begin{cases}
\sum_{i=0}^{p-1} M_r(u-i) \theta_i & x \in (x_0,x_0+L), \\
0 & \text{o.w.}
\end{cases}
\end{equation}
  For periodic splines, $x$ is circularly mapped into the correct range, and arguments of $B$ and $M$ as well as indices to $\theta$ are understood as modulo $p$.
%whereas for aperiodic splines all $\theta_j$ outside the range $j\in[0,p-1]$ are considered to be zero.

  Next, we note that $M_r(u)$ is nonzero only in the range $u\in(0,r)$, so the only necessary values of $i$ in the above sum are $i\in(u-r,u) \cap \mathbb Z = \{\floor{u}-r+1, \floor{u}-r+2, \ldots, \floor{u}\}$.  Defining $d=u-\floor{u}$, and using the fact that this evaluation only requires parameters $\{\theta\}_{\floor{u}-r+1}^{\floor{u}} \equiv {\vec \theta}_{\floor{u}}$, we have (for any $u\in (\Delta,L/h+\Delta)$)
\begin{align}
\label{eq:Bu}
B(u;\theta) &= \sum_{i=\floor{u}-r+1}^{\floor{u}} M_r(u-i) \theta_i = \sum_{i=0}^{r-1} M_r(i+1-d) \theta_{i+\floor{u}-r+1} \notag \\
 &\equiv P_r(d;\mathbf M)^T \cdot {\vec \theta}_{\floor{u}}
   = {\vec d_r}^T \cdot \mathbf M \cdot {\vec \theta}_{\floor{u}}
.
\end{align}

  Defining the above sum as a dot product with $(r-1)^{\text{th}}$ order polynomial functions $P_r(d; \mathbf M)^T = [P_r(d; M^0), \ldots, P_r(d; M^{r-1})]$ ($M^j$ denotes a column of the $r \times r$ matrix $\mathbf M$) emphasizes the linear nature of the spline basis functions.  To further compress the notation, define a vector of powers ${\vec d}_r^T = [d^0,d^1,\ldots,d^{r-1}]$ so that $P_r(d;c) = {\vec d}_r^T \cdot c$ and $P_r(d;\mathbf M)^T = {\vec d}_r^T \cdot \mathbf M$.

  We can solve for the coefficients $M^j$ in terms of the standard polynomial coefficients for $M_r(u) = P_r(u;c^{\floor{u}})$ by using the Binomial theorem to expand $M_r(i+1-d)$.
\begin{align}
\label{eq:binom}
P_r(i+a d;c) &= \sum_{j=0}^{r-1} c_j (i+a d)^j = \sum_{j=0}^{r-1} \sum_{k=0}^j c_j \binom{j}{k} i^{j-k} a^k d^k \\
	&\equiv \sum_{k=0}^{r-1} b_k d^k = {\vec d}_r^T \cdot b \notag \\
	&\Rightarrow b_k = [\mathbf B(a,i) \cdot c]_k \notag
\end{align}
The last expression defines a lower-triangular matrix of binomial coefficients,
\begin{equation}
[\mathbf B(a,i)]_{jk} = \begin{cases}
\binom{j}{k} i^{j-k} a^k & k \le j, \\
0 & \text{otherwise.}
\end{cases}
\end{equation}

  Comparing the terms in Eqs.~\ref{eq:Bu} and \ref{eq:binom}, the columns of $\mathbf M$ are given by:
\begin{equation}
\label{eq:Mcoef}
M^j = \mathbf B(-1,j+1) \cdot c^j
\end{equation}
Where $c^j$ are the polynomial coefficients used for normal B-spline interpolation on an interval $j$.  A further simplification of \eqref{eq:Mcoef} is possible by noting that $P_r(u;c^{\floor{u}}) = M_r(u) = M_r(r-u) = P_r(r-u;c^{\floor{r-u}})$ so that any of the columns in \eqref{eq:Mcoef} can be equivalently expressed as:
\begin{equation}
M^j = \mathbf B(+1,r-1-j) \cdot c^{r-1-j}
\end{equation}
specifically, replacing the last half of the columns ($j=\ceil{r/2},\cdots,r-1$) only requires the initial computation of $\ceil{r/2}$ coefficient vectors, $\{c\}_0^{\ceil{r/2}}-1$.

  We can also note because of linearity that
\begin{equation}
B^{(n)}(u;\theta) = {\vec \theta}_{\floor{u}}^T \cdot \mathbf M^T \cdot \mathbf D_n \cdot {\vec d}_{r-n}
.
\end{equation}
Which represents differentation using an $r \times (r-n)$ matrix $\mathbf D_n$ with nonzero entries $[\mathbf D_n]_{ij}=i!/(i-n)!$ only on the diagonal $i-j=n$.

  Finally, we can assemble the integral
%\begin{align}
%\label{eq:Eint}
%E &= V^{-1} \kappa \int_{x_0}^{x_0+L} g^{(n)}(u(x);\theta)^2 x^k dx \\
% &= V^{-1} \kappa h^{1+k-2n} \int_{u(x_0)}^{u(x_0+L)} g^{(n)}(u;\theta)^2 (u+%\Delta+x_0/h)^k du
%\end{align}
\begin{equation}
\label{eq:Eint}
E = V^{-1} \kappa h^{1+k-2n} \int_{u_0}^{u_1}
	g^{(n)}(u;\theta)^2 (u+\Delta+x_0/h)^k du
\end{equation}
where we have restricted ourselves to the case where $\rho(x) T(x)/T_0 = \kappa x^k$ for illustration and transformed $x$ to $u$ and $(x_0,x_0+L)$ to $(u_0,u_1)$ as above.  Now we decompose the integral into separate contributions from each interval in $\{I\} = \{(u_0,\floor{u_0}+1), (\floor{u_0}+1,\floor{u_0}+2), \cdots, (\floor{u_1},u_1)\}$ to give a sum of integrals from each interval identified by its unique $\floor{u}$.
\begin{align}
\label{eq:Qdecomp}
\int_{u_0}^{u_1} g^{(n)}(&u;\theta)^2 (u+\Delta+x_0/h)^k du \notag \\
 &= \sum_{\{I\}} \int_I g^{(n)}(u;\theta)^2 (u+\Delta+x_0/h)^k du \notag \\
 &= \sum_{\{I\}} \int_I {\vec \theta}_I^T \cdot \mathbf M^T
        \cdot \mathbf D_n   \cdot {\vec d}_{r-n} {\vec d}_{r-n}^T
        \cdot \mathbf D_n^T \cdot \mathbf M \cdot {\vec \theta}_I
  (u+\Delta+x_0/h)^k du \notag \\
 &\equiv \sum_{\{I\}} {\vec \theta}_I^T \cdot \mathbf Q_I \cdot {\vec \theta}_I
\end{align}
Assuming we have a general form for the integral of $\rho$ times a power of $x$ on each interval, we write
\begin{equation}
\label{eq:Qi}
\mathbf Q_I = \mathbf M^T \cdot \mathbf D_n \cdot \mathbf F \cdot \mathbf D_n^T \cdot \mathbf M
,
\end{equation}
with $[\mathbf F]_{ij} = \int_I d^{i+j} (d+\floor{u}+\Delta+x_0/h)^k dd$, $i,j = 0,1,\ldots,r-n-1$.

  The complete penalty matrix $\mathbf Q$ is then assembled according to \eqref{eq:Qdecomp} by shifting each of the matrices \eqref{eq:Qi} into its proper location (multiplying ${\vec \theta}_I = \{\theta\}_{\floor{u}-r+1}^{\floor{u}}$).  For periodic splines, $\rho$ must also be periodic and the edges of this matrix are wrapped to correspond to the correct parameters.  For aperiodic splines, edges are simply discarded, since their corresponding parameters have been assumed to be zero.

\section*{ Availability}

  A python implementation of the methods used in this report have been made available as a part of the ForceSolve project on sourceforge.net \\
(\href{http://forcesolve.sourceforge.net}{http://forcesolve.sourceforge.net}).
This software was created as a proof-of-concept for a general method to fit molecular dynamics data to arbitrary stochastic integration models.

\section*{ Acknowledgments}

We thank Donald French and Randall Laviolette for helpful discussions, and
gratefully acknowledge the Army MURI program (DAAD19-02-1-0227), NSF grant CHE-0709560, and the DOE Computational Science Graduate Fellowship (DE-FG02-97ER25308) for the support of this work.

\bibliographystyle{acmtrans-ims}
\bibliography{stsplines}

\end{document}